\documentclass[11pt]{article}
\usepackage{eacl2017}
\usepackage{times}
\usepackage{url}
\usepackage{latexsym}

\usepackage{caption}
\usepackage{subcaption}
\usepackage{enumitem}
\usepackage{mathtools}
\usepackage{tabularx}
\usepackage{threeparttable}
\usepackage{tikz}

\usetikzlibrary{positioning,shapes,shadows,arrows,automata,chains,fit}
\usepackage[linesnumbered,ruled,vlined]{algorithm2e}
\eaclfinalcopy 
\def\eaclpaperid{10} 

\graphicspath{ {/}}

\title{Transition-Based Deep Input Linearization}

\author{Ratish Puduppully \dag \thanks{\hspace{1.5mm}Part of the work was done when the author was a visiting student at Singapore University of Technology and Design.} , Yue Zhang \ddag, Manish Shrivastava \dag\\
  \dag Kohli Center on Intelligent Systems (KCIS),\\
International Institute of Information Technology, Hyderabad (IIIT Hyderabad)\\
  \ddag Singapore University of Technology and Design \\
  ratish.surendran@research.iiit.ac.in   \hfill yue\_zhang@sutd.edu.sg \\ m.shrivastava@iiit.ac.in\\}

\date{}

\begin{document}
\maketitle
\begin{abstract}
Traditional methods for deep NLG adopt pipeline approaches comprising stages such as constructing syntactic input, predicting function words, linearizing the syntactic input and generating the surface forms. Though easier to visualize, pipeline approaches suffer from error propagation. In addition, information available across modules cannot be leveraged by all modules. We construct a transition-based model to jointly perform linearization, function word prediction and morphological generation, which considerably improves upon the accuracy compared to a pipelined baseline system. On a standard deep input linearization shared task, our system achieves the best results reported so far.
\end{abstract}

\section{Introduction}
Natural language generation (NLG) \cite{reiter1997building,white2004reining} aims to synthesize natural language text given input syntactic, semantic or logical representations. It has been shown useful in various tasks in NLP, including machine translation \cite{P07-1002,zhang-EtAl:2014:EMNLP}, abstractive summarization \cite{barzilay2005sentence} and grammatical error correction \cite{lee2006automatic}. 
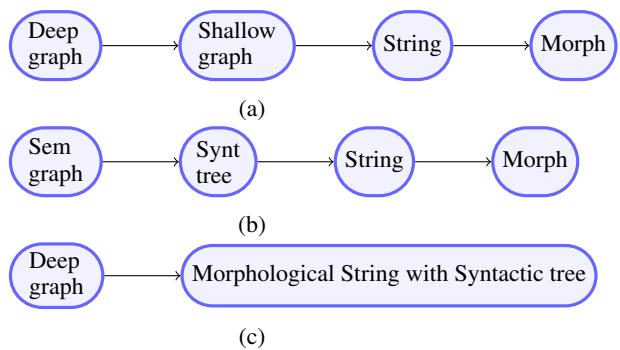
\begin{figure}[t]
\small
\begin{subfigure}[b]{0.4\textwidth}
\begin{tikzpicture}[
roundnode/.style={circle, draw=green!60, fill=green!5, very thick, minimum size=6mm},
squarednode/.style={rounded rectangle, draw=blue!60, fill=blue!5, very thick, minimum size=9mm, align=left},
]
\node[squarednode]      (sem)                              {Deep\\graph};
\node[squarednode]      (graph)       [right=of sem] {Shallow\\graph};
\node[squarednode]      (string)       [right=of graph] {String};
\node[squarednode]      (morph)      [right=of string] {Morph};

\draw[->] (sem.east) -- (graph.west);
\draw[->] (graph.east) -- (string.west);
\draw[->] (string.east) -- (morph.west);
\end{tikzpicture}
\caption{}
\label{fig-pipeline-b}
\end{subfigure}

\begin{subfigure}[b]{0.4\textwidth}
\begin{tikzpicture}[
roundnode/.style={circle, draw=green!60, fill=green!5, very thick, minimum size=6mm},
squarednode/.style={rounded rectangle, draw=blue!60, fill=blue!5, very thick, minimum size=9mm, align=left},
]
\node[squarednode]      (sem)                              {Sem\\graph};
\node[squarednode]      (synt)       [right=of sem] {Synt\\tree};
\node[squarednode]      (string)       [right=of synt] {String};
\node[squarednode]      (morph)      [right=of string] {Morph};

\draw[->] (sem.east) -- (synt.west);
\draw[->] (synt.east) -- (string.west);
\draw[->] (string.east) -- (morph.west);
\end{tikzpicture}
\caption{}
\label{fig-pipeline-a}
\end{subfigure}

\begin{subfigure}[b]{0.4\textwidth}
\begin{tikzpicture}[
roundnode/.style={circle, draw=green!60, fill=green!5, very thick, minimum size=6mm},
squarednode/.style={rounded rectangle, draw=blue!60, fill=blue!5, very thick, minimum size=8mm, align=left},
]
\node[squarednode]      (sem)                              {Deep\\graph};
\node[squarednode]      (synt)       [right=of sem] {Morphological String with Syntactic tree};

\draw[->] (sem.east) -- (synt.west);

\end{tikzpicture}
\caption{}
\label{fig-pipeline-c}
\end{subfigure}
\caption{Linearization pipelines (a) NLG pipeline with deep input graph, (b) pipeline based on the meaning text theory, (c) this paper.} 
\label{pipeline}
\vspace*{-1em}
\end{figure}

A line of traditional methods treat the problem as a pipeline of several independent steps \cite{bohnet2010broad,E09-1097,W00-1401,W00-0306,C98-1112}. For example, shown in Figure \ref{fig-pipeline-a}, a pipeline based on the meaning text theory (MTT) \cite{melʹvcuk1988dependency} splits NLG into three independent steps 1. syntactic generation: generating an unordered and lemma-formed syntactic tree from a semantic graph, introducing function words; 2. syntactic linearization: linearizing the unordered syntactic tree; 3. morphological generation: generating the inflection for each lemma in the string.

\begin{figure}[t]
\centering
\small
\includegraphics[width=0.48\textwidth]{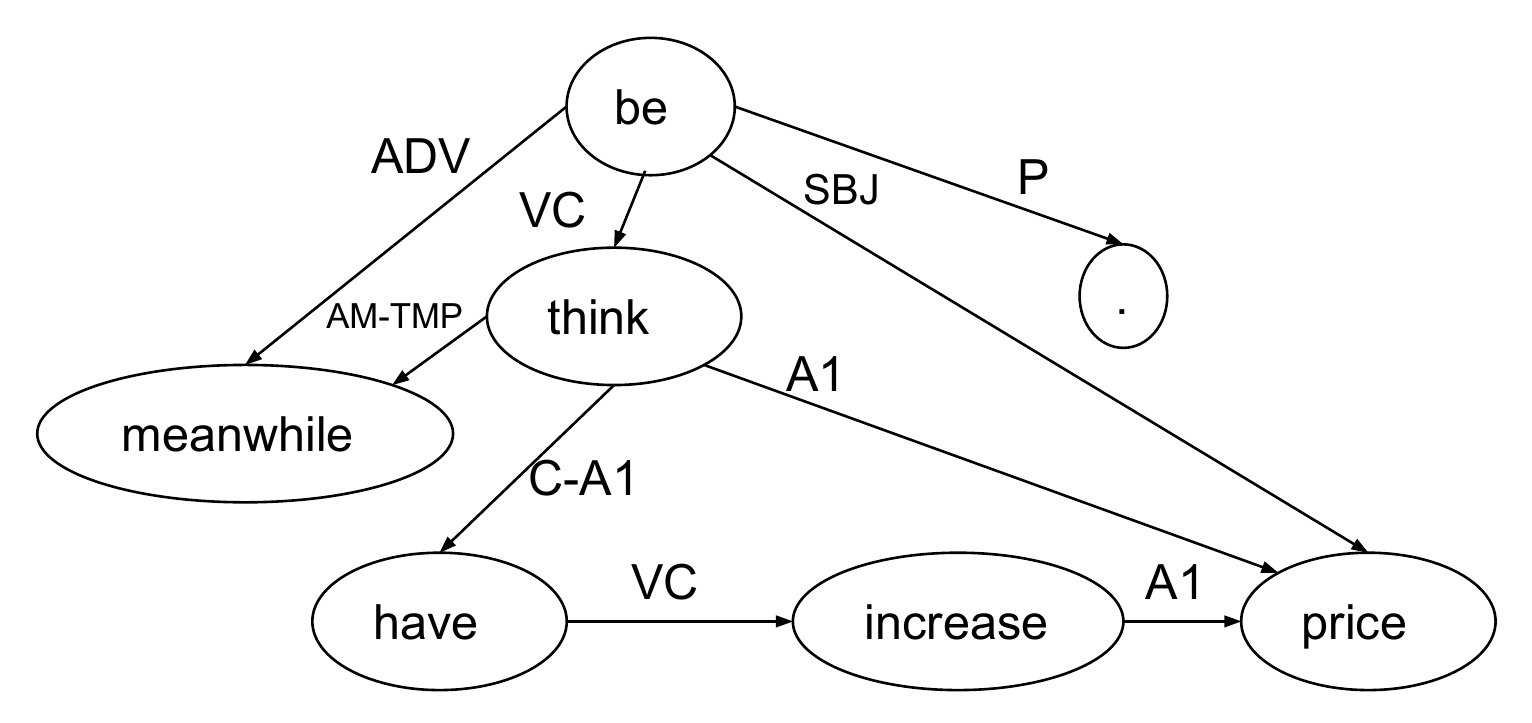}
\caption{Sample deep graph for the sentence: {\it meanwhile, prices are thought to have increased.} Note that words are replaced by their lemmas. The function word {\it to} and comma are absent in graph.}
\label{fig-deep-graph}
\vspace*{-1em}
\end{figure}

In this paper we focus on deep graph as input. Exemplified in Figure \ref{fig-deep-graph}, the deep input type is intended to be an abstract representation of the meaning of a sentence. Unlike semantic input, where the nodes are semantic representations of input, deep input is more surface centric, with lemmas for each word being connected by semantic labels \cite{W13-2322,mel2015semantics}. In contrast to shallow syntactic trees, function words in surface forms are not included in deep graphs \cite{belz2011first}. Deep inputs can more commonly occur as input of NLG systems where entities and content words are available, and one has to generate a grammatical sentence using them with only provision for inflections of words and introduction of function words. Such usecases include summarization, dialog generation etc.

A pipeline of deep input linearization is shown in Figure \ref{fig-pipeline-b}. Generation involves predicting the correct word order, deciding inflections and also filling in function words at the appropriate positions. The worst-case complexity is $n!$ for permuting {\it n} words, $2^n$ for function word prediction (assuming that a function word can be inserted after each content word), and $2^n$ for inflection generation (assuming two morphological forms for each lemma). On the dataset from the First Surface Realisation Shared Task, \newcite{bohnet2011stumaba} achieved the best reported results on linearizing deep input representation, following the pipeline of Figure \ref{fig-pipeline-a} (with input as deep graph instead of semantic graph). They construct a syntactic tree from deep input graph followed by function word prediction, linearization and morphological generation. A rich set of features are used at each stage of the pipeline and for each adjacent pair of stages, an SVM decoder is defined. 

Pipelined systems suffer from the problem of error propagation. In addition, because the steps are independent of each other, information available in a later stage is not made use of in the earlier stages. We introduce a transition-based \cite{J08-4003} method for {\it joint} deep input surface realisation integrating linearization, function word prediction and morphological generation. The model is shown in Fig \ref{fig-pipeline-c}, as compared with the pipelined baseline in Fig \ref{fig-pipeline-b}. For a directly comparable baseline, we construct a pipeline system of function words prediction, linearization and morphological generation similar to the pipeline of \newcite{bohnet2011stumaba}, but with the following difference. Our baseline pipeline system makes function word prediction for a deep input graph, whereas \newcite{bohnet2011stumaba} have a preprocessing step to construct a syntactic tree from the deep input graph, which is given as input to the function word prediction module. Our pipeline is directly comparable to the joint system with regard to the use of information.

Standard evaluations show that: 1. Our joint model for deep input surface realisation achieves significantly better scores over its pipeline counterpart. 2. We achieve the best results reported on the task. Our system scores 1 BLEU point better over \newcite{bohnet2011stumaba} without using any external resources. We make the source code available at \url{https://github.com/SUTDNLP/ZGen/releases/tag/v0.3}.


\section{Related Work}
Related work can be broadly summarized into three areas: abstract word ordering, applications of meaning-text theory and joint modelling of  NLP tasks. In abstract word ordering \cite{E09-1097,zhang2013partial,zhang2015discriminative}, \newcite{de2014word} compose phrases over individual words and permute the phrases to achieve linearization. \newcite{schmaltz2016word} show that strong surface-level language models are more effective than models trained with syntactic information for the task of linearization. Transition-based techniques have also been explored \cite{LiuZCQ15,liu-zhang:2015:EMNLP,N16-1058}. To our knowledge, we are the first to use transition-based techniques for {\it deep} input linearization.

There has been work done in the area of sentence linearization using meaning-text theory \cite{melʹvcuk1988dependency}. \newcite{belz2011first} organized a shared task on both shallow and deep linearization according to meaning-text theory, which provides a standard benchmark for system comparison. \newcite{song2014joint} achieved the best results for the task of {\it shallow}-syntactic linearization.
Using SVM models with rich features, \newcite{bohnet2011stumaba} achieved state-of-art results on the task of {\it deep} realization. While they built a pipeline system, we show that joint models can be used to overcome limitations of the pipeline approach giving the best results. 

Joint models for NLP have shown effectiveness in recent years. Though having to tackle increased search space, they overcome issues with error propagation in pipelined models. Joint models have been explored for grammar-based approaches to surface realisation using HPSG and CCG \cite{I05-1015,W06-1661,P08-1022,D09-1043,white2006efficient,carroll1999efficient}. Joint models have been proposed for word segmentation and POS-tagging \cite{zhang2010fast}, POS-tagging and syntactic chunking \cite{sutton2007dynamic},  segmentation and normalization \cite{tao-zhang-zhang-ji:2015:EMNLP}, syntactic linearization and morphologization \cite{song2014joint}, parsing and NER \cite{N09-1037}, entity and relation extraction \cite{P14-1038} and so on. We propose a first joint model for deep realization, integrating linearization, function word prediction and morphological generation.

\section{Baseline}
We build a baseline following the pipeline in Figure \ref{fig-pipeline-b}. Three stages are involved: 1. prediction of function words, inserting the predicted function words in the deep graph, resulting in a {\it shallow} graph; 2. linearizing the shallow graph; 3. generating the inflection for each lemma in the string.
\subsection{Function Word Prediction}
\label{function-word-prediction}

\begin{figure}[t]
\centering
\small
\includegraphics[width=0.48\textwidth]{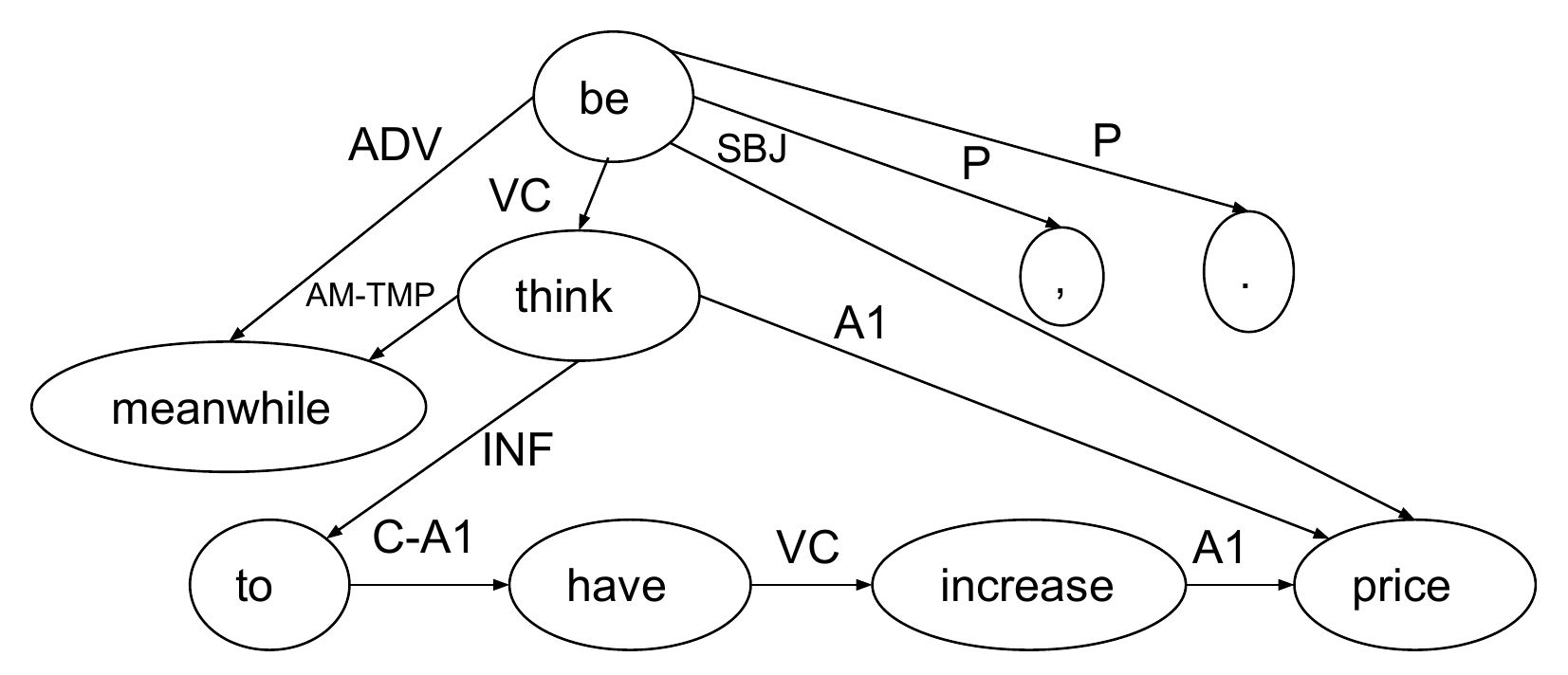}
\caption{Equivalent shallow graph for Figure \ref{fig-deep-graph}.}
\label{fig-shallow-graph}
\vspace*{-1em}
\end{figure}

In the First Surface Realisation Shared Task dataset \cite{belz2011first}, there are three classes of function words to predict: {\it to} infinitive, {\it that} complementizer and {\it comma}. We implement classifiers to predict these classes of function words locally at respective positions in the deep graph resulting in a shallow graph (Figure \ref{fig-shallow-graph}). At each location the input  is a node and output is a class indicating if {\it to} or {\it that} need to inserted under the node or the count of {\it comma} to be introduced under the node. 

Table \ref{function-word-to-that} shows the feature templates for classification of {\it to} infinitives and {\it that} complementizers and Table \ref{function-word-comma} shows the feature templates for predicting the count of {\it comma} child nodes for each non-leaf node in the graph. These feature templates are a subset of features used in the joint model (Section \ref{joint-method}) with the exceptions being word order features, which are not available here for the pipeline system, since earlier stages cannot leverage features in subsequent outcomes. We use averaged perceptron classifier \cite{W02-1001} to predict function words, which is consistent with the joint model. 
\begin{table}
\centering
\footnotesize
\begin{tabular}{|l|}

  \hline
  Features for predicting function words including \\{\it to} infinitive, {\it that} complementizer \\
\hline
	WORD({\it n}); POS({\it n});	WORD({\it c}) \\
  \hline
  
\end{tabular}
\caption{Feature templates for the prediction of function words- {\it to} infinitive and {\it that} complementizer. Indices on the surface string: {\it n} -- word index; {\it c} -- child of {\it n}; Functions: WORD -- word at index; POS -- part-of-speech at index.}
\label{function-word-to-that}
\end{table}

\begin{table}
\centering
\footnotesize
\begin{tabular}{|l|}

  \hline
  Features for predicting count of {\it comma} \\
\hline
	WORD({\it n}); POS({\it n})\\
	BAG(WORD-MOD({\it n}))\\
	BAG(LABEL-MOD({\it n}))\\
  \hline
  
\end{tabular}
\caption{Feature templates for the comma prediction system. Indices on the surface string: {\it n} -- word index; Functions: WORD -- word at index; POS -- part-of-speech at index; WORD-MOD -- modifiers of index; LABEL-MOD -- dependency labels of modifiers; BAG -- set. }
\label{function-word-comma}
\vspace*{-1em}
\end{table}

\subsection{Linearization}
\label{linearization}
The next step is linearizing the graph, which we solve using a novel transition-based algorithm.
\subsubsection{Transition-Based Tree Linearization} 
\label{transition-based-linearization}
\newcite{LiuZCQ15} introduce a transition-based model for tree linearization. The approach extends from transition-based parsers \cite{nivre2004deterministic,chen2014fast}, where {\it state} consists of {\it stack} to hold partially built outputs and a {\it queue} to hold input sequence of words. In case of linearization, the input is a set of words. Liu et al. therefore use a set to hold the input instead of a queue. State is represented by a tuple {\it ($\sigma$, $\rho$, A)}, where $\sigma$ is stack to store partial derivations, $\rho$ is set of input words and {\it A} is the set of dependency relations that have been built. There are three transition actions:	
\begin{itemize}[noitemsep,nolistsep]
\item {\sc Shift}-{\it Word}-POS -- shifts {\it Word} from $\rho$, assigns POS to it and pushes it to top of stack as {\it S}$_0$;
\item {\sc LeftArc}-LABEL -- constructs dependency arc {\it S}$_1 \xleftarrow{LABEL}$ {\it S}$_0$ and pops out second element from top of stack {\it S}$_1$;
\item {\sc RightArc}-LABEL -- constructs dependency arc {\it S}$_1 \xrightarrow{LABEL}$ {\it S}$_0$ and pops out top of stack {\it S}$_0$.
\end{itemize} 
The sequence of actions to linearize the set \{{\it he, goes, home}\} is {\sc Shift}-{\it he}, {\sc Shift}-{\it goes}, {\sc Shift}-{\it home}, {\sc RightArc}-OBJ, {\sc LeftArc}-SBJ.

The full set of feature templates are shown in Table 2 of \newcite{LiuZCQ15}, partly shown in Table \ref{tab:features-templates-standard}. The features include word({\it w}), POS({\it p}) and dependency label ({\it l}) of elements on stack and their descendants {\it S}$_0$, {\it S}$_1$, {\it S}$_{0,l}$, {\it S}$_{0,r}$ etc. For example, word on top of stack is {\it S}$_0${\it w} and word on first left child of {\it S}$_0$ is {\it S}$_{0,l}${\it w}. These are called configuration features. They are combined with all possible actions to score the action.
\newcite{N16-1058} extend \newcite{LiuZCQ15} by redefining features to address feature sparsity and introduce lookahead features, thereby achieving highest accuracies on task of abstract word ordering. 

\subsubsection{Shallow Graph Linearization}
Our transition based graph linearization system extends from \newcite{N16-1058}. In our case, the input is a shallow graph instead of a syntactic tree, and hence the search space is larger. On the other hand, the same set of actions can still be applied, with additional constraints on valid actions given each configuration (Section \ref{possible-transition-actions}). Table \ref{tbl:arc-standard-shallow-linearization} shows the sequence of transition actions to linearize shallow graph in Figure \ref{fig-shallow-graph}. 

\begin{table}[t]
\centering
\scriptsize
\begin{tablenotes}
\item \textbf{Input lemmas}: \{think$_1$, price$_2$, .$_3$, increase$_4$,  be$_5$, have$_6$, meanwhile$_7$, ,$_8$, to$_9$\} 
\end{tablenotes}
\vspace*{0.5em}
\begin{tabular}{llp{1.5cm}p{1cm}l}
\hline
  & Transition     & $\sigma$    & $\rho$     & A \\
0 &                & []          & \{1...7\} & $\emptyset$                 \\ 
1 & {\sc SH}-meanwhile    & [7]         & \{1...6,8,9\} &                             \\ 
2 & {\sc SH}-,    & [7 8]         & \{1...6,9\} &                             \\ 
3 & {\sc SH}-price    & [7 8 2]       & \{1,3,4,5,6,9\} &                             \\ 
4 & {\sc SH}-be    & [7 8 2 5]     & \{1,3,4,6,9\} &                             \\ 
5 & {\sc SH}-think    & [7 8 2 5 1]   & \{3,4,6,9\}   &                             \\ 
6 & {\sc SH}-to    & [7 8 2 5 1 9]   & \{3,4,6\}   &                             \\ 
7 & {\sc SH}-have    & [7 8 2 5 1 9 6]   & \{3,4\}   &                             \\ 
8 & {\sc SH}-increase    & [7 8 2 5 1 9 6 4]   & \{3\}   &                             \\ 
9 & {\sc RA} & [7 8 2 5 1 9 6]   & \{3\}    & $A \cup \{6\rightarrow 4\}$ \\ 
10 & {\sc RA} & [7 8 2 5 1 9]   & \{3\}     & $A \cup \{9\rightarrow 6\}$ \\ 
11 & {\sc RA} & [7 8 2 5 1]     & \{3\}     & $A \cup \{1\rightarrow 9\}$ \\ 
12 & {\sc RA} & [7 8 2 5]       & \{3\}     & $A \cup \{5\rightarrow 1\}$ \\ 
13 & {\sc SH}-.    & [7 8 2 5 3]     & \{\}      &                             \\ 
14& {\sc RA} & [7 8 2 5]       & \{\}      & $A \cup \{5\rightarrow 3\}$ \\ 
15& {\sc LA}  & [7 8 5]         & \{\}      & $A \cup \{2\leftarrow 5\}$  \\ 
16& {\sc LA}  & [7 5]         & \{\}      & $A \cup \{8\leftarrow 5\}$  \\ 
17& {\sc LA}  & [5]         & \{\}      & $A \cup \{7\leftarrow 5\}$  \\ 

\hline
\end{tabular}
\caption{Transition action sequence for linearizing the graph in Figure \ref{fig-shallow-graph}. SH  - {\sc Shift}, RA - {\sc RightArc}, LA - {\sc LeftArc}. POS is not shown in \textsc{Shift} actions.}
\label{tbl:arc-standard-shallow-linearization}
\vspace*{-0.5em}
\end{table}	
  
\subsubsection{Obtaining Possible Transition Actions Given a Configuration}

\begin{figure}[t]
\includegraphics[width=0.48\textwidth]{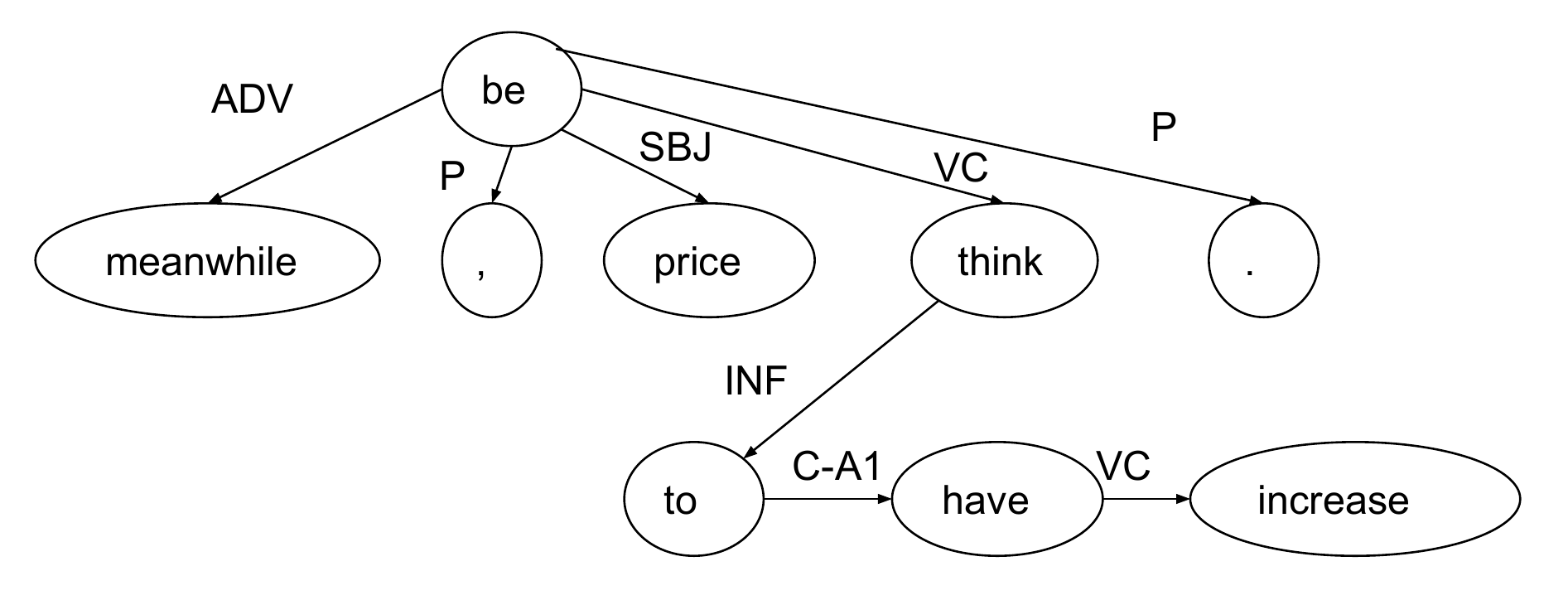}
\caption{Equivalent syntactic tree for Figure \ref{fig-deep-graph}.}
\label{fig:shallow-tree}
\end{figure}

\label{possible-transition-actions}
\begin{algorithm}
\small
\DontPrintSemicolon 
\KwIn{A state $s=([\sigma|j\ i], \rho, A)$ and input graph $C$}
\KwOut{A set of possible transition actions $T$}
$T \gets \emptyset$\;
\If{$s.\sigma == \emptyset$} {
  \For{$k \in s.\rho$} {
    $T \gets T \cup {(\textsc{Shift}, POS, k)}$\;
  }
} \Else {

  \If{$\exists k, k \in (\textsc{DirectChildren}(i) \cap s.\rho)$} {
   		\textsc{ShiftSubtree}$(i,\rho)$
  } \Else {
	\If{A.\textsc{LeftChild}(i) is \textsc{Nil}}{
   		\textsc{ShiftSubtree}$(i,\rho)$
  }  
	\If{$\{j \rightarrow i\} \in C \land $ A.\textsc{LeftChild}(j) is \textsc{Nil}}{
      $T \gets T \cup {(\textsc{RightArc})}$\;
			\If{$i \in $ \textsc{Descendant}$(j)$}{
				\textsc{ProcessDescendant($i,j$)}			
			}      		
			\If{$i \in $ \textsc{Sibling}$(j)$}{			
				\textsc{ProcessSibling($i,j$)}			
			}      					      
    } \ElseIf{$\{j \leftarrow i\} \in C$} {
      $T \gets T \cup {(\textsc{LeftArc})}$\;
			\If{$i \in $ \textsc{Sibling}$(j)$}{			
				\textsc{ProcessSibling($i,j$)}			
			}      					      
      }\Else{
      	\If{$size(s.\sigma)$  == 1}{
			\textsc{ShiftParentAndSiblings($i$)}
      	}\Else{
			\If{$i \in $ \textsc{Descendant}$(j)$}{
				\textsc{ProcessDescendant($i,j$)}			
			}      		
			\If{$i \in $ \textsc{Sibling}$(j)$}{			
				\textsc{ProcessSibling($i,j$)}			
			}      					
      	}      
      }
	}	
	  
  }

\Return{$T$}\;
\caption{{\sc GetPossibleActions} for shallow graph linearization}
\label{algo:possible_action_graph}
\end{algorithm}

\begin{algorithm}
\small
\DontPrintSemicolon 
\KwIn{A state \textit{s=([$\sigma|j\ i$], $\rho$, A), input\_node and graph C.}}
\KwOut{\textit{DC} direct child nodes of input node }
	\textit{DC $\gets \emptyset$}\;
  \For{\textit{k $\in$ (\textsc{C.Children}(input\_node))}}{
	\textit{Parents $\gets$ {\textsc{C.Parents}(k)}} \\
	\If{\textit{Parents.size} == 1}{
		\textit{DC $\gets$ DC $\cup$ k	}
	}\Else{
		\For{\textit{m $\in$ Parents}}{
			\If{\textsc{A.LeftChild}(m) is not \textsc{Nil} $\lor$ m == input\_node}{
				continue
			}
			\If{\textit{m $\cap$ s.$\rho$}}{
				goto \textit{OutsideLoop}
			}
			
			\If{\textit{m $\in \sigma \land$ \textsc{$\sigma$.IsAncestor}(m,C)}}{
				goto \textit{OutsideLoop}
			}
		}
		\textit{DC $\gets$ DC $\cup$ k}
	}
	\textit{OutsideLoop:}	
  }
  return \textit{DC}
\caption{\textsc{DirectChildren}}
\label{algo:direct_children}
\end{algorithm}

\begin{algorithm}
\small
\DontPrintSemicolon 
\KwIn{A state $s=([\sigma|j\ i], \rho, A)$, graph $C$, head $k$}
\KwOut{a set of possible Transition actions $T$}
{\it T $\gets \emptyset$}\\
{\it T $\gets$ T $\cup$ {(\textsc{Shift}, POS, k)}}\\
{\it queue q}\\
{\it q.push(k)}\\
\While {{\it q is not empty}}{
{\it front = q.pop()}\\
\For{\it m $\in$ (\textsc{C.Children}(front) $\cap$ s.$\rho$)} {
{\it q.push(m)} \\
{\it T $\gets$ T $\cup$ {(\textsc{Shift}, POS, m)}}
}
}
\caption{\textsc{ShiftSubtree}}
\label{algo:ShiftSubtree}
\end{algorithm}

The purpose of a {\sc GetPossibleActions} function is to find out the set of transition actions that can lead to a valid output given a certain state. This is because not all sequences of actions correspond to  a well-formed output. Essentially, given a state $s=([\sigma| j\ i], \rho, A)$ and an input graph {\it C}, the Decoder extracts syntactic tree from the graph (cf. Figure \ref{fig:shallow-tree} extracted from Figure \ref{fig-shallow-graph}), outputting \textsc{RightArc}, \textsc{LeftArc} only if the corresponding arc exists in {\it C}.  The corresponding pseudocode is shown in Algorithm \ref{algo:possible_action_graph}.

In particular, if node {\it i} has {\it direct child nodes} in {\it C}, the descendants of {\it i} are shifted (line 6-7) (see Algorithm \ref{algo:ShiftSubtree}). Here {\it direct child nodes} (see Algorithm \ref{algo:direct_children}) include those child nodes of {\it i} for which {\it i} is the only parent or if there is more than one parent then every other parent is shifted on to the stack without possibility to reduce the child node. If no {\it direct child node} is in the buffer, then all graph descendants of {\it i} are shifted. Now, there are three configurations possible between {\it i} and {\it j}: 1. {\it i} and {\it j} are directly connected in {\it C}. This results in \textsc{RightArc} or \textsc{LeftArc} action; 2. {\it i} is descendant of {\it j}. In this case the parents of {\it i} (such that they are descendants of {\it j}) and siblings of {\it i} through such parents are shifted. 3. {\it i} is sibling of {\it j}. In this case, parents of {\it i} and their descendants are shifted such that {\it A} remains consistent. Because the input is a graph, more than one of the above configuration can occur simultaneously. More detailed discussion related to {\sc GetPossibleActions} is given in Appendix \ref{sec:appendix-B}.

\subsubsection{Feature Templates}
There are three sets of features. The first is the set of baseline linearization feature templates from Table 2 in \newcite{LiuZCQ15}, partly shown in Table \ref{tab:features-templates-standard}. The second is a set of {\it lookahead features} similar to that of \newcite{N16-1058}, shown in Table  \ref{tab:features-templates}.\footnote{Here $L_{cls}$ represents set of arc labels of child nodes (of word to shift {\it L}) shifted on the stack, $L_{clns}$ represents set of arc labels of child nodes not shifted on the stack, $L_{cps}$ the POS set of shifted child nodes, $L_{cpns}$ the POS set of unshifted child nodes, $L_{sls}$ the set of arc labels of shifted siblings, $L_{slns}$ the set of arc labels of unshifted siblings, $L_{sps}$ the POS set of shifted siblings, $L_{cpns}$ the POS set of unshifted siblings, $L_{pls}$ the set of arc labels of shifted parents, $L_{plns}$ the set of arc labels of unshifted parents, $L_{pps}$ the POS set of shifted parents, $L_{ppns}$ the POS set of unshifted parents.} Parent lookahead feature in \newcite{N16-1058} is defined for the only parent. For graph linearization, however, the parent lookahead feature need to be defined for set of parents. The third set of features in Table \ref{tab:new-features-templates} are newly introduced for Graph Linearization. {\it Arc$_{\it left}$} is a binary feature indicating if there is left arc between $S_0$ and $S_1$, whereas {\it Arc$_{\it right}$} indicates if there is a right arc.  {\it L$_{\it is\_descendant}$} is a binary feature indicating if $L$ is descendant of $S_0$, and {\it L$_{\it is\_parent\_or\_sibling}$} indicates if it is a parent or sibling of $S_0$. {\it $S_{\it 0descendants\_shifted}$} is binary feature indicating if all the descendants of $S_0$ are shifted. 

Not having POS in the input dataset, we compute the feature templates for POS making use of the most frequent POS of the lemma in the gold training data. For the features with dependency labels, we use the input graph labels. 

\begin{table}[t]
\centering
\footnotesize
\begin{tabularx}{.45\textwidth}{X}

  \hline
  Unigrams \\

  \hline
  $S_0w$; $S_0 p$; $S_{0,l} w$;  $S_{0,l} p$;  $S_{0,l} l$;   $S_{0,r} w$;  $S_{0,r} p$;  $S_{0,r} l$;  \\

  \hline \hline
  Bigram \\
  \hline
  $S_{0} w S_{0,l} w$; $S_{0} w S_{0,l} p$; $S_{0} w S_{0,l} l$; $S_{0} p S_{0,l} w$;\\ 
  \hline \hline
  Linearization \\
  \hline
  $ w_0$; $p_0$; $w_{-1}w_0$; $p_{-1}p_0$; $w_{-2}w_{-1}w_0$; $p_{-2}p_{-1}p_0$\\
\hline

\end{tabularx}
\caption{Baseline linearization feature templates. A subset is shown here. For the full feature set, refer to Table 2 of \newcite{LiuZCQ15}.}
\label{tab:features-templates-standard}
\end{table}

\begin{table}[t]
\centering
\footnotesize
\begin{tabularx}{.45\textwidth}{X}
\hline 
set of label and POS of child nodes of $L$\\ 
\hline 
$L_{cls};L_{clns};L_{cps};L_{cpns};$\\
$S_0wL_{cls};S_0pL_{cls};S_1wL_{cls};S_1pL_{cls};$\\
\hline 
\hline  
set of label and POS of first-level siblings of $L$\\ 
\hline 
$L_{sls};L_{slns};L_{sps};L_{spns};$\\
$S_0wL_{sls};S_0pL_{sls};S_1wL_{sls};S_1pL_{sls};$\\
\hline 
\hline  
set of label and POS of parents of $L$\\ 
\hline 
$L_{pls};L_{plns};L_{pps};L_{ppns};$\\
$S_0wL_{pls};S_0pL_{pls};S_1wL_{pls};S_1pL_{pls};$\\
\hline 

\end{tabularx}
\caption{Lookahead linearization feature templates for the word {\it L} to shift. A subset is shown here. For the full feature set, refer to Table 2 of \newcite{N16-1058}. An identical set of feature templates are defined for {\it S}$_0$.}
\label{tab:features-templates}
\end{table}

\begin{table}
\centering
\footnotesize
\begin{tabularx}{.45\textwidth}{X}
\hline 
arc features between $S_0$ and $S_1$ \\
\hline
{\it Arc$_{\it{left}}$}; {\it Arc$_{\it{right}}$}; \\
\hline
\hline
lookahead features for {\it L} \\
\hline
$L_{\it is\_descendant}$; $L_{\it is\_parent\_or\_sibling}$; \\
\hline
\hline
are all descendants of {\it S$_0$} shifted\\
\hline
$S_{\it{0descendants\_shifted}}$;\\
\hline
\hline
feature combination \\
\hline
$S_{\it{0descendants\_shifted}}${\it Arc}$_{\it left}$; \\
$S_{\it{0descendants\_shifted}}${\it Arc}$_{\it{right}}$;\\
$S_{\it{0descendants\_shifted}}$ $L_{\it{is\_descendant}}$; \\
$S_{\it{0descendants\_shifted}}$ $L_{\it{is\_parent\_or\_sibling}}$; \\
\hline 

\end{tabularx}
\caption{Graph linearization feature templates}
\label{tab:new-features-templates}
\end{table}

\subsubsection{Search and Learning}
  We follow \newcite{N16-1058} and \newcite{LiuZCQ15}, applying the learning and search framework of \newcite{zhang2011syntactic}. Pseudocode is shown in Algorithm \ref{algo:beam-search}. It performs beam search holding {\it k} best states in an agenda at each incremental step. At the start of decoding, agenda holds the initial state. At a step, for each state in the agenda, each of transition actions in {\sc GetPossibleActions} is applied. The top-{\it k} states are updated in the agenda for the next step. The process repeats for 2{\it n} steps as each word needs to be shifted once on to the stack and reduced once. After 2{\it n} steps, the highest scoring state in agenda is taken as the output. The complexity of algorithm is {\it n}$^2$, as it takes 2{\it n} steps to complete and during each step, the number of transition actions is proportional to $\rho$.
  Given a configuration $C$, the score of a possible action $a$ is calculated as: \vspace*{-0.3em}
  \[Score(a)=\vec{\theta} \cdot \vec{\Phi(C,a)} ,\vspace*{-0.3em}\]
  where $\vec{\theta}$ is the model parameter vector and $\vec{\Phi(C,a)}$ denotes a feature vector consisting of {\it configuration} and {\it action} components. Given a set of labeled training examples, the averaged perceptron with early update \cite{collins2004incremental} is used.

\begin{algorithm}[t]
\small
\DontPrintSemicolon 
\KwIn{$C$, a set of input syntactic constraints}
\KwOut{The highest-scored final state}
candidates $\gets ([\ ], set(1..n), \emptyset)$ \;
agenda $\gets \emptyset$ \;
\For{$i \gets 1..2n$} {
  \For{$s$ \textbf{in} candidates} {
    \For {action \textbf{in} \textsc{GetPossibleActions}($s$, $C$)} {
      agenda $\gets$ {\sc Apply}($s$, action)\;
    }
  }
  candidates $\gets$ {\sc Top-K}(agenda) \;
  agenda $\gets \emptyset$ \;
}
{\it best} $\gets$ {\sc Best}(candidates) \;
\Return {\it best}\;
\caption{transition-based linearization}
\label{algo:beam-search}
\end{algorithm}  
  
\subsection{Morphological Generation}
\label{morphological-generation}
The last step is to inflate the lemmas in the sentence. There are three POS categories, including nouns, verbs and articles, for which we need to generate morphological forms. We use Wiktionary\footnote{https://en.wiktionary.org/} as a basis and write a small set of rules similar to that used in \newcite{song2014joint}, listed in Table \ref{tbl-morphology-generation}, to generate a candidate set of inflections. An averaged perceptron classifier \cite{W02-1001} is trained for each lemma. For distinguishing between singular and plural candidate verb forms, the feature templates in Table \ref{sing-plural-forms} are used.

\begin{table}[t]
\centering
\footnotesize
\begin{tabularx}{.45\textwidth}{|X|}
\hline 
\textbf{Rules for {\it be}} \\ 
\hline 
attr[`partic'] == `pres' $\rightarrow$ being\\ 
attr[`partic'] == `past' $\rightarrow$ been\\ 
attr[`tense'] == `past'\\ 
$\>\>$sbj.attr[`num'] == `sg' $\rightarrow$ was\\ 
$\>\>$sbj.attr[`num'] == `pl' $\rightarrow$ were\\ 
$\>\>$other $\rightarrow$ [was,were]\\ 
attr[`tense'] == `pres'\\ 
$\>\>$sbj.attr[`num'] == `sg' $\rightarrow$ is\\ 
$\>\>$sbj.attr[`num'] == `pl' $\rightarrow$ are\\ 
$\>\>$other $\rightarrow$ [am,is,are]  \\ 
\hline 
\end{tabularx} 

\begin{tabularx}{.45\textwidth}{|X|}
\hline 
\textbf{Rules for other verbs} \\ 
\hline 
attr[`partic'] == `pres' $\rightarrow$ wik.get(lemma, VBG) \\
attr[`partic'] == `past' $\rightarrow$ wik.get(lemma, VBN ) \\
attr[`tense'] == `past' $\rightarrow$ wik.get(lemma, VBD) \\
attr[`tense'] == `pres' \\
$\>\>$sbj.attr[`num'] == `sg' $\rightarrow$ wik.get(lemma, VBZ )\\
$\>\>$other $\rightarrow$ wik.getall(lemma) \\
\hline 
\end{tabularx} 

\begin{tabularx}{.45\textwidth}{|X|}
\hline 
\textbf{Rules for other types} \\ 
\hline 
lemma==a $\rightarrow$ [a,an] \\
lemma==not $\rightarrow$ [not,n't] \\
attr[`num'] == `sg' $\rightarrow$ wik.get(lemma,NNP/NN) \\ 
attr[`num'] == `pl' $\rightarrow$ wik.get(lemma,NNPS/NNS) \\ 
\hline 
\end{tabularx} 
\caption{Lemma rules. All rules are in the format: conditions $\rightarrow$ candidate inflections. Nested conditions are listed in multi-lines with indentation. {\it wik} denotes english wiktionary.}
\label{tbl-morphology-generation}
\end{table}

\begin{table}
\centering
\footnotesize
\begin{tabular}{|l|}

  \hline
  Features for predicting singular/ plural verb forms\\
\hline
	WORD({\it n-1})WORD({\it n-2})WORD({\it n-3}); COUNT\_SUBJ({\it n});\\COUNT({\it n-1})COUNT({\it n-2})COUNT({\it n-3}); SUBJ({\it n});\\
	WORD({\it n-1})WORD({\it n-2}); COUNT({\it n-1})COUNT({\it n-2});\\
	WORD({\it n-1}); COUNT({\it n-1});
	WORD({\it n+1}); COUNT({\it n+1});\\	
  \hline
\end{tabular}
\caption{Feature templates for predicting singular/ plural verb forms. Indices on the surface string: {\it n} -- word index; Functions: WORD -- word at index {\it n}; COUNT -- word at {\it n} is singular or plural form; SUBJ -- word at subject of {\it n}; COUNT\_SUBJ -- word at subject of {\it n} is singular or plural form.}
\label{sing-plural-forms}
\end{table}

\section{Joint Method}
\label{joint-method}
We design a joint method for function word prediction (Section \ref{function-word-prediction}), linearization (Section \ref{linearization}) and morphological generation (Section \ref{morphological-generation}) by further extending the transition-based system of Section \ref{linearization}, integrating actions for function word prediction and morphological generation.

\subsection{Transition Actions}
\begin{figure}[t]
\centering
\small
\includegraphics[width=0.48\textwidth]{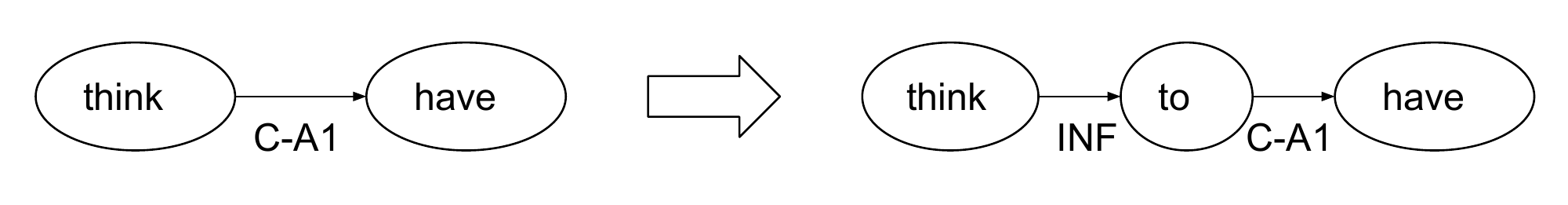}
\caption{Example for {\sc SplitArc}-{\it to}.}\label{fig:split-arc}
\vspace*{-1em}
\end{figure}

In addition to \textsc{Shift}, \textsc{LeftArc} and \textsc{RightArc} in Section \ref{transition-based-linearization}, we use the following new transition actions for inserting function words:
\begin{itemize}[noitemsep,nolistsep]
\item {\sc Insert}, inserts comma at the present position;
\item {\sc SplitArc}-{\it Word}, splits an arc in the input graph {\it C}, inserting a function word between the words connected by the arc. Here {\it Word} specifies the function word being inserted (Figure \ref{fig:split-arc}).
\end{itemize}

We generate a candidate set of inflections for each lemma following the approach in Section \ref{morphological-generation}. For each candidate inflection, we generate a corresponding {\sc Shift} transition action. The rules in Table \ref{tbl-morphology-generation} are used to prune impossible inflections.\footnote{For example in Figure \ref{fig-deep-graph}, {\it price} is the subject of {\it be} and if {\it be} is in present tense and {\it price} is in plural form, the inflections \{{\it am, is, was, were}\} are impossible and {\it are} is the correct inflection for {\it be}. We therefore generate transition action as {\sc Shift}-{\it are}.}

Table \ref{tbl:arc-standard-linearization} shows the transition actions to linearize the graph in Figure \ref{fig-deep-graph}. These newly introduced transition actions result in variability in the number of transition actions. With function word prediction, the number of transition actions for a bag of {\it n} words is not necessarily {\it 2n-1}. For example, considering an {\sc Insert}, {\it {\sc SplitArc}-to} or {\it {\sc  SplitArc}-that} action post each {\sc Shift} action, the maximum number of possible actions is {\it 5n-1}. This variance in the number of actions can impact the linear separability of state items. Following \newcite{zhu2013fast}, we use {\sc Idle} actions as a form of padding method, which results in completed state items being further expanded up to {\it 5n-1} steps. The joint model uses the same perceptron training algorithm and similar features compared to the baseline model.

\begin{table}[t]
\centering
\scriptsize
\begin{tablenotes}
\item \textbf{Input lemmas}: \{think$_1$, price$_2$, .$_3$, increase$_4$,  be$_5$, have$_6$, meanwhile$_7$\} 
\end{tablenotes}
\vspace*{0.5em}
\begin{tabular}{lllll}
\hline
  & Transition     & $\sigma$    & $\rho$     & A \\
0 &                & []          & \{1...7\} & $\emptyset$                 \\ 
1 & {\sc SH}-meanwhile    & [7]         & \{1...6\} &                             \\ 
2 & {\sc IN}    & [7]         & \{1...6\} &                             \\ 
3 & {\sc SH}-prices    & [7 2]       & \{1,3,4,5,6\} &                             \\ 
4 & {\sc SH}-are    & [7 2 5]     & \{1,3,4,6\} &                             \\ 
5 & {\sc SH}-thought    & [7 2 5 1]   & \{3,4,6\}   &                             \\ 
6 & {\sc SP}-to    & [7 2 5 1] & \{3,4,6\}     &                       \\ 
7 & {\sc SH}-have    & [7 2 5 1 6]   & \{3,4\}   &                             \\ 
8 & {\sc SH}-increased    & [7 2 5 1 6 4]   & \{3\}   &                             \\ 
9 & {\sc RA} & [7 2 5 1 6]   & \{3\}     & $A \cup \{6\rightarrow 4\}$ \\ 
10 & {\sc RA} & [7 2 5 1]     & \{3\}     & $A \cup \{1\rightarrow 6\}$ \\ 
11 & {\sc RA} & [7 2 5]       & \{3\}     & $A \cup \{5\rightarrow 1\}$ \\ 
12 & {\sc SH}-.    & [7 2 5 3]     & \{\}      &                             \\ 
13& {\sc RA} & [7 2 5]       & \{\}      & $A \cup \{5\rightarrow 3\}$ \\ 
14& {\sc LA}  & [7 5]         & \{\}      & $A \cup \{2\leftarrow 5\}$  \\ 
15& {\sc LA}  & [5]         & \{\}      & $A \cup \{7\leftarrow 5\}$  \\ 

\hline
\end{tabular}
\caption{Transition action sequence for linearizing the sentence in Figure \ref{fig-deep-graph}. SH  - {\sc Shift}, SP - {\sc SplitArc}, RA - {\sc RightArc}, LA - {\sc LeftArc}, IN - {\sc Insert}. POS is not shown in \textsc{Shift} actions.}
\label{tbl:arc-standard-linearization}
\vspace*{-0.5em}
\end{table}	
  
\subsection{Obtaining Possible Transition Actions Given a Configuration}
Given a state $s=([\sigma| j\ i], \rho, A)$ and an input graph {\it C}, the possible transition actions include as a subset the transition actions in Algorithm \ref{algo:possible_action_graph} for shallow graph linearization. In addition, for each lemma being shifted, we enumerate its inflections and create {\sc Shift} transition actions for each inflection. Further, we predict {\sc SplitArc}, {\sc Insert} and {\sc Idle} actions to handle function words. If node {\it i} has a child node in {\it C}, which is not shifted, we predict {\sc SplitArc} and {\sc Insert}. If {\it i} is sibling to {\it j}, we predict {\sc Insert}. If both the stack and buffer are empty, we predict {\sc Idle}. Pseudocode for {\sc GetPossibleActions} for the joint method is shown in Algorithm \ref{algo:possible_action_deep_graph}. 
\begin{algorithm}[!ht]
\small
\DontPrintSemicolon 
\KwIn{A state $s=([\sigma|j\ i], \rho, A)$ and graph $C$}
\KwOut{A set of possible transition actions $T$}
$T \gets \emptyset$\;
\If{$s.\sigma == \emptyset$} {
  \For{$k \in s.\rho$} {
    $T \gets T \cup {(\textsc{Shift}, POS, k)}$\;
  }
} \Else {

  \If{$\exists k, k \in (\textsc{DirectChildren}(i) \cap s.\rho)$} {
   		\textsc{ShiftSubtree}$(i,\rho)$
  } \Else {
	\If{A.\textsc{LeftChild}(i) is \textsc{Nil}}{
   		\textsc{ShiftSubtree}$(i,\rho)$
  }  
	\If{$\{j \rightarrow i\} \in C \land $ A.\textsc{LeftChild}(j) is \textsc{Nil}}{
      $T \gets T \cup {(\textsc{RightArc})}$\;
			\If{$i \in $ \textsc{Descendant}$(j)$}{
				\textsc{ProcessDescendant($i,j$)}			
			}      		
			\If{$i \in $ \textsc{Sibling}$(j)$}{			
				\textsc{ProcessSibling($i,j$)}			
			}      					      
    } \ElseIf{$\{j \leftarrow i\} \in C$} {
      $T \gets T \cup {(\textsc{LeftArc})}$\;
			\If{$i \in $ \textsc{Sibling}$(j)$}{			
				\textsc{ProcessSibling($i,j$)}			
			}      					      
      }\Else{
      	\If{$size(s.\sigma)$  == 1}{
			\textsc{ShiftParentAndSiblings($i$)}
      	}\Else{
			\If{$i \in $ \textsc{Descendant}$(j)$}{
				\textsc{ProcessDescendant($i,j$)}			
			}      		
			\If{$i \in $ \textsc{Sibling}$(j)$}{			
				\textsc{ProcessSibling($i,j$)}			
			}      					
      	}      
      }
	}	
	  
  }

\If{{\it C.Children(i) }$\wedge s.\rho  \neq \emptyset $}{
$T \gets T \cup {(\textsc{SplitArc}-to)}$ \\
$T \gets T \cup {(\textsc{SplitArc}-that)}$ \\
}
\If{{\it C.Children(i) }$\wedge s.\rho  \neq \emptyset \lor i \in $ \textsc{Sibling}(j) }{
$T \gets T \cup {(\textsc{Insert})}$ \\
}
\If{\textsc{$s.\sigma== \emptyset$ $\wedge$ $s.\rho== \emptyset$}}{
 $T \gets T \cup {(\textsc{Idle})}$\;
}
\Return{$T$}\;
\caption{{\sc GetPossibleActions} for deep graph linearization, where $C$ is a input graph}
\label{algo:possible_action_deep_graph}
\end{algorithm}

\section{Experiments}
\subsection{Dataset} 
We work on the deep dataset from the Surface Realisation Shared Task \cite{belz2011first}\footnote{http://www.nltg.brighton.ac.uk/research/sr-task/}. Sentences are represented as sets of unordered nodes with labeled semantic edges between them. Semantic representation is obtained by merging Nombank \cite{L04-1228}, Propbank \cite{J05-1004} and syntactic dependencies. Edge labeling follows PropBank annotation scheme such as \{{\it A0, A1, ... An\}}. The nodes are annotated with lemma and where appropriate number, tense and participle features. Function words including {\it that} complementizer, {\it to} infinitive and commas are omitted from  the input. There are two punctuation features for information about brackets and quotes. Table \ref{training-instance} shows a sample training instance. 

\begin{table}
\scriptsize
\begin{tablenotes}
\item \textbf{Input} (unordered lemma-formed graph):
\end{tablenotes}
\begin{tabular}{lcclll}
\textbf{Sem} & \textbf{ID} & \textbf{PID} & \textbf{Lemma} & \textbf{Attr} & \textbf{Lexeme}\\
SROOT &	1 & 0 &	be &	tense=pres	&	are\\
ADV	&	2	&	1	&	meanwhile	&	& meanwhile\\
P	&	3	&	1	&	.	&	& .\\
SBJ	&	4	&	1	&	start.02	&	num=pl	& starts\\
A1	&	5	&	4	&	housing	&	num=sg	& housing\\
AM-TMP	&	6	&	4	&	september	&	num=sg	& september\\
VC	&	9	&	1	&	think.01	&	partic=past	& thought\\
A1	&	 4	&	9& & &\\
C-A1	&	10	&	9	&	have	&&have\\
VC	&	11	&	10	&	inch.01	&	partic=past &inched\\
A1	&	 4	&	11 & & &\\
A5	&	12	&	11	&	upward&&upward\\
\end{tabular} 
\caption{Deep type training instance from Surface Realisation Shared Task 2011. {\it Sem} -- semantic label, {\it ID} -- unique ID of node within graph, {\it PID} -- the ID of the parent, {\it Attr} -- Attributes such as partic (participle), tense or number, {\it Lexeme} -- lexeme which is resolved using wiktionary and rules in Table \ref{tbl-morphology-generation}.}
\label{training-instance}
\vspace*{-1.5em}
\end{table}

Out of 39k total training instances, 2.8k are non-projective, which we discard. We exclude instances which result in non-projective dependencies mainly because our transition actions predict only projective dependencies. It has been derived from the arc-standard system \cite{J08-4003}. There are 1.8k training instances with a mismatch between edges in the input deep graph and {\it gold output tree}. The gold output tree is the corresponding shallow tree from the shared task. We approach the task of linearization as extracting a linearized tree from the input semantic graph. So we exclude those instances which do not have edges corresponding to gold tree i.e mismatch between edges of gold tree and input graph. After excluding these instances, we have 34.3k training instances. We also exclude 800 training instances where the function words {\it to} and {\it that} have more than one child, and around 100 training instances where function words' parent and child nodes are not connected by an arc in the deep graph. The above cases are deemed annotation mistakes. We thus train on a final subset of 33.4k training instances. The development set comprises 1034 instances and the test set comprises 2398 instances.  Evaluation is done using the BLEU metric \cite{P02-1040}. 

\section{Development Results}
\subsection{Influence of Beam Size}
\begin{figure}[t]
\centering
\small
\includegraphics[width=0.43\textwidth]{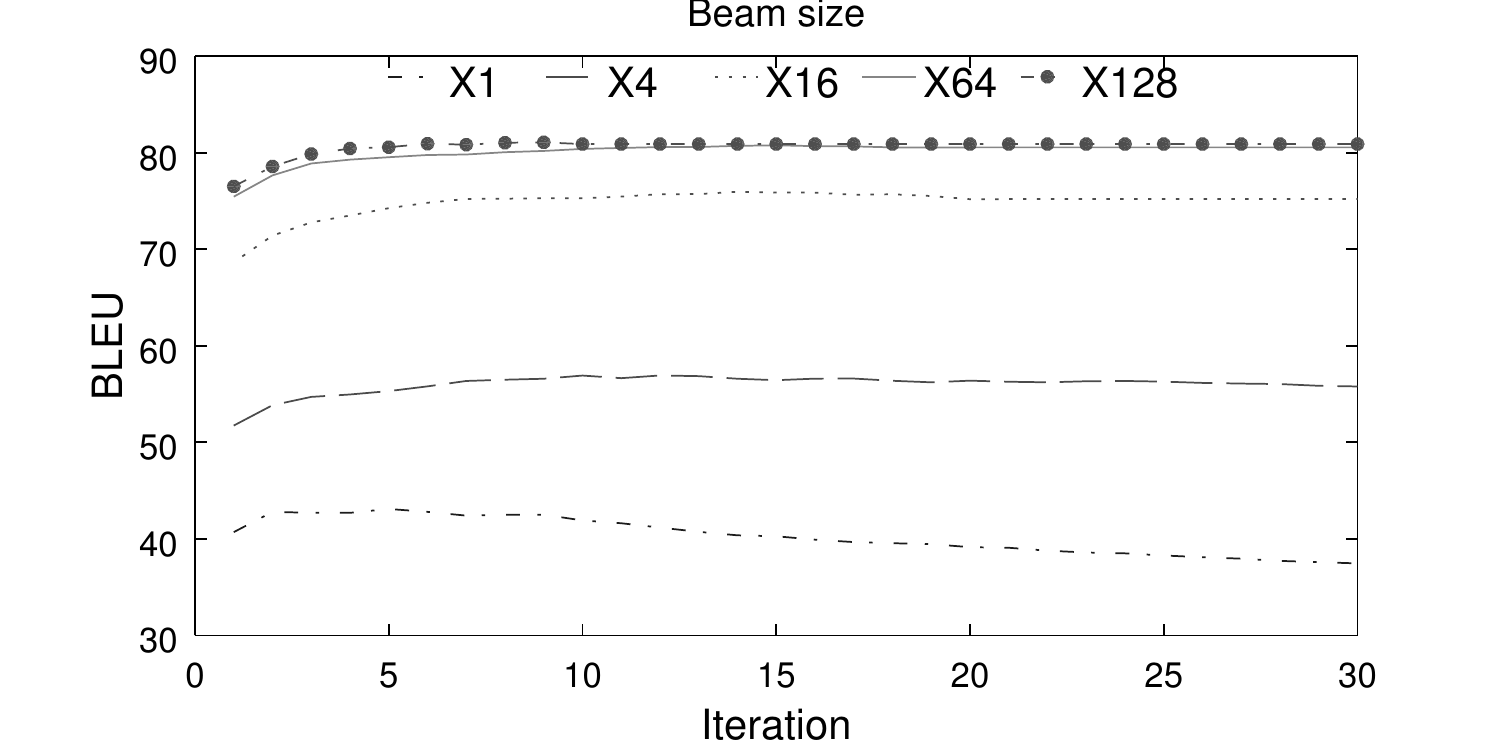}
\caption{Influence of beam sizes.}\label{fig:dev-res-beam}
\end{figure}

We study the effect of beam size on the accuracies of joint model in Figure \ref{fig:dev-res-beam}, by varying the beam size and comparing the accuracies on development dataset over training iterations. Beam sizes of 64 and 128 perform the best. However, beam size 128 does not improve the performance significantly, yet is twice as slow compared to a beam size 64. So we retain a 64 beam  for further experiments.  

\subsection{Pipeline vs Joint Model}

We compare the results of the joint model with the pipeline baseline system. Table \ref{tbl-dev-function-word} shows the development results of function word prediction, and Table \ref{tbl:development_results} shows the overall development results. Our joint model of Transition-Based Deep Input Linearization (TBDIL) achieves an improvement of 5 BLEU points over the pipeline using the same feature source and training algorithm. Thanks to the sharing of word order information, the joint model improves function word prediction compared to the pipeline, which forbids such feature integration because function word prediction is the first step, taken before order becomes available.

\begin{table}[t]
\centering
\small
\begin{tabular}{|l|c|c|}
\hline 
• & Pipeline & Joint \\ 
\hline 
{\it to} infinitive & 92.7 & 94.1\\ 
\hline 
{\it that} complementizer & 70.6 & 76.5\\ 
\hline 
count of {\it comma} & 60.2 & 63.3 \\ 
\hline 
\end{tabular} 
\caption{Average F-measure for function word prediction for development set.}
\label{tbl-dev-function-word}
\end{table}

\begin{table}[t]
\centering
\small
\begin{tabular}{|c|c|}
\hline 
System & BLEU Score \\ 
\hline 
Pipeline &  75.86\\ 
\hline 
TBDIL &\textbf{80.77} \\ 
\hline 
\end{tabular} 
\caption{Development results.}
\label{tbl:development_results}
\end{table}
\section{Final Results}
Table \ref{tbl:test_results} shows the final results. The best performing system for the Shared Task was STUMABA-D by \newcite{bohnet2011stumaba}, which leverages a large-scale n-gram language model. The joint model TBDIL significantly outperforms the pipeline system and achieves an improvement of 1 BLEU point over STUMABA-D, obtaining 80.49 BLEU without making use of external resources.

\begin{table}[t]
\centering
\small
\begin{tabular}{|c|c|}
\hline 
System & BLEU Score \\ 
\hline 
STUMABA-D & 79.43 \\ 
\hline 
Pipeline & 70.99  \\ 
\hline 
TBDIL & \textbf{80.49} \\ 
\hline 
\end{tabular} 
\caption{Test results.}
\label{tbl:test_results}
\end{table}
\section{Analysis}

\begin{table}[!t]
\setlength{\tabcolsep}{3.3pt}
\centering
\footnotesize
\begin{tabular}{l||p{0.36\textwidth}}
\hline
     & output   \\
\hline
ref. & if it does n't yield on these matters and eventually begin talking directly to the anc\\
Pipeline & if it does not to yield on these matters and eventually begin talking directly to the anc\\
TBDIL & if it does n't yield on these matters and eventually begin talking directly to the anc \\
\hline
ref. & economists who read september 's low level of factory job growth as a sign of a slowdown \\
Pipeline & september 's low level of factory job growth who as a sign of a slowdown reads economists \\
TBDIL & economists who read september 's low level of factory job growth as a sign of a slowdown\\
\hline
\end{tabular}
\caption{Example outputs.}
\label{tab:example-outputs}
\vspace*{-1em}
\end{table}

Table \ref{tab:example-outputs} shows sample outputs from the Pipeline system and the corresponding output from TBDIL. In the first instance, the function word {\it to} is incorrectly predicted in the arc between nodes {\it does} and {\it yield} in the pipeline system. In case of TBDIL, the n-gram feature helps avoid incorrect insertion of {\it to} which demonstrates the advantage of integrating information across stages. In the second instance, because of incorrect linearization, there is error propagation to morphological generation in the pipeline system. In particular, {\it economists} is linearized to the object part of the sentence and the subject is singular. This, in turn, results in the incorrect prediction of morphological form of verb {\it read} as its singular variant. In TBDIL, in contrast, the joint modelling of linearization and morphology helps ordering the sentence correctly.

\section{Conclusion}
We showed the usefulness of a joint model 
for the task of Deep Linearization, by taking \cite{N16-1058} as the baseline and extending it to perform joint graph linearization, function word prediction and morphological generation. To our knowledge, this is the first work to use Transition-Based method for joint NLG from semantic structure. 
Our system gave the highest scores reported for the NLG 2011 shared task on Deep Input Linearization \cite{belz2011first}. 


\section*{Acknowledgments}
We thank Litton Kurisinkel for helpful discussions and the anonymous reviewers for their detailed and constructive comments. Yue Zhang is supported by the Singapore Ministry of Education (MOE) AcRF Tier 2 grant T2MOE201301.

\bibliography{eacl2017}
\bibliographystyle{eacl2017}
\newpage
\appendix

\section{Obtaining possible transition actions given a configuration for Shallow Graph}
\label{sec:appendix-B}
During shallow linearization, a state is represented by $s=([\sigma| j\ i], \rho, A)$ and {\it C} is the input graph. Given {\it C}, the Decoder outputs actions which extract syntactic tree from the graph. Thus the Decoder outputs \textsc{RightArc} or \textsc{LeftArc} only if corresponding arc exists in {\it C}. The detailed pseudocode is given in Algorithm \ref{algo:possible_action_graph}. If {\it i} has {\it direct child nodes} in {\it C}, the descendants of {\it i} are shifted (line 6-7) (see Algorithm \ref{algo:ShiftSubtree}). Here, {\it direct child nodes} (see Algorithm \ref{algo:direct_children}) include those child nodes of {\it i} for which {\it i} is the only parent or if there is more than one parent then every other parent is shifted on to the stack without possibility to reduce the child node. If no {\it direct child node} is in buffer, then descendants of {\it i} are shifted (line 9-10). Now, there are three configurations possible between {\it i} and {\it j}: 1. {\it i} and {\it j} are connected by arc in {\it C}. This results in \textsc{RightArc} or \textsc{LeftArc} action; 2. {\it i} is descendant of {\it j}. In this case the parents of {\it i} (such that they are descendants of {\it j}) and siblings of {\it i} through such parents are shifted. 3. {\it i} is sibling of {\it j}. In this case, the parents of {\it i} and their descendants are shifted such that {\it A} remains consistent. Additionally, because the input is a graph structure, more than one of the above configuration can occur simultaneously. We analyse the three configurations in detail below.

Since the {\it direct child nodes} of {\it i} are shifted, $\{j \leftarrow i\}$ results in a \textsc{LeftArc} action (line 18). Also because the input is a graph, {\it i} can be a sibling node of {\it j}. In this case, the valid parents and siblings of {\it i} are shifted. We iterate through the other elements in stack to identify the valid parents and siblings. These conditions are encapsulated in \textsc{ProcessSibling} (line 20). Conditions for \textsc{RightArc} are similar to that of \textsc{LeftArc} with the following differences. We ensure that there is no left arc relationship for {\it j} in {\it A} (line 11). If there is a left arc relationship for {\it j} in {\it A}, it means that in an arc-standard setting, the \textsc{RightArc} actions for {\it j} have already been made. If {\it i} is a descendant of {\it j}, valid parents and siblings of {\it i} are shifted. We iterate through the parents of {\it i} and those parents which are in turn descendants of {\it j} and not shifted on to the stack are valid parents. We shift the parent and the subtree through each such parent. These conditions are denoted by \textsc{ProcessDescendant} (line 14). 

If there is no arc between {\it j} and {\it i} and there is only one element on the stack, then the parents and siblings of {\it i} are shifted (line 22-23). If there is more than one element on the stack, and if {\it i} is descendant of {\it j}, then we use \textsc{ProcessDescendant} (line 25-26). If {\it i} is sibling to {\it j} we use \textsc{ProcessSibling} (line 27-28).

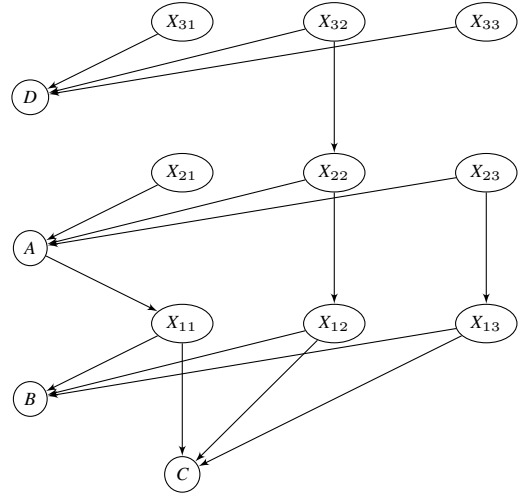
\begin{figure}[t]
\centering
\scriptsize
\begin{tikzpicture}

\tikzset{vertex/.style = {shape=ellipse,draw,minimum size=1.5em}}
\tikzset{edge/.style = {->,> = latex'}}
\node[vertex] (C) at (2,0) {{\it C}};
\node[vertex] (B) at (0,1) {{\it B}};
\node[vertex] (X11) at (2,2) {{\it X$_{11}$}};
\node[vertex] (X12) at (4,2) {{\it X$_{12}$}};
\node[vertex] (X13) at (6,2) {{\it X$_{13}$}};
\node[vertex] (A) at (0,3) {{\it A}};
\node[vertex] (X21) at (2,4) {{\it X$_{21}$}};
\node[vertex] (X22) at (4,4) {{\it X$_{22}$}};
\node[vertex] (X23) at (6,4) {{\it X$_{23}$}};
\node[vertex] (D) at (0,5) {{\it D}};
\node[vertex] (X31) at (2,6) {{\it X$_{31}$}};
\node[vertex] (X32) at (4,6) {{\it X$_{32}$}};
\node[vertex] (X33) at (6,6) {{\it X$_{33}$}};

\draw[edge] (X11) to (B);
\draw[edge] (X12) to (B);
\draw[edge] (X13) to (B);
\draw[edge] (X11) to (C);
\draw[edge] (X12) to (C);
\draw[edge] (X13) to (C);
\draw[edge] (X21) to (A);
\draw[edge] (X22) to (A);
\draw[edge] (X23) to (A);
\draw[edge] (X31) to (D);
\draw[edge] (X32) to (D);
\draw[edge] (X33) to (D);
\draw[edge] (A) to (X11);
\draw[edge] (X22) to (X12);
\draw[edge] (X32) to (X22);
\draw[edge] (X23) to (X13);
\end{tikzpicture}
\caption{Sample graph to illustrate {\sc ProcessSibling}}
\label{fig-process-sibling}
\end{figure}

Consider an example to see the working of {\sc ProcessSibling} in detail. In {\sc ProcessSibling}, we need to ensure that {\it i} is in stack because of sibling relation with {\it j} and we need to shift the valid parent nodes of i and their descendants. We call these valid nodes {\it inflection points}. Consider the following stack entries [{\it D, A, B, C}] with {\it C} as stack top. Assume that the input graph is as in Figure \ref{fig-process-sibling}. {\it C} is sibling of {\it B} through {\it B}'s parents {\it X$_{11}$, X$_{12}$, X$_{13}$}. Out of these, only {\it X$_{11}$} and {\it X$_{12}$} are valid parents.  {\it X$_{13}$} is sibling to {\it A} through {\it A}'s parent {\it X$_{23}$}. But {\it X$_{23}$} is in turn neither descendant of {\it D} nor sibling of {\it D}. Thus {\it X$_{13}$} is not a valid inflection point for {C}. Now, {\it X$_{12}$} is sibling of {\it A} through {\it A}'s parent {\it X$_{22}$. X$_{22}$} is in turn sibling of {\it D} through {\it X$_{32}$}. Thus there is a path to the stack bottom through a path of siblings/ descendant. In case of {\it X$_{11}$}, {\it X$_{11}$} is descendant of stack element {\it A} and is thus valid. {\it X$_{11}$} and {\it X$_{12}$} are called valid inflection points. If inflection point is a common parent to both {\it S$_0$} and {\it S$_1$} then inflection point and its descendants are shifted. Instead, if inflection point is ancestor to {\it S$_0$}, then parents of {\it S$_0$} (say {\it P$_0$}) which are descendants of inflection point are shifted. Additionally, descendants of {\it P$_0$} are shifted.

\end{document}